\documentclass[twoside,leqno,twocolumn]{article}
\usepackage{ltexpprt}
\usepackage{algorithm,algpseudocode}
\usepackage{amsfonts}
\usepackage{amsmath}
\usepackage{algorithm}
\usepackage{graphics}
\usepackage{multirow}
\begin{document}

\title{\Large Low-rank Label Propagation for \\Semi-supervised Learning with 100 Millions Samples}
\author{Raphael Petegrosso\thanks{Department of Computer Science and Engineering, University of Minnesota Twin Cities, USA.}
\thanks{co-first authors}
\\
\and
Wei Zhang\footnotemark[1] \footnotemark[2] \\
\and
Zhuliu Li\footnotemark[1] \\
\and
Yousef Saad\footnotemark[1] \\
\and
Rui Kuang\footnotemark[1] \thanks{kuang@cs.umn.edu}}
\date{}

\maketitle


\begin{abstract} \small\baselineskip=9pt
The success of semi-supervised learning crucially relies on the scalability to a huge amount of unlabelled data that are needed to capture the underlying manifold structure for better classification. Since computing the pairwise similarity between the training data is prohibitively expensive in most kinds of input data, currently, there is no general ready-to-use semi-supervised learning method/tool available for learning with tens of millions or more data points. In this paper, we adopted the idea of two low-rank label propagation algorithms, GLNP (Global Linear Neighborhood Propagation) and Kernel Nystr\"{o}m Approximation, and implemented the parallelized version of the two algorithms accelerated with Nesterov's accelerated projected gradient descent for Big-data Label Propagation (BigLP).
The parallel algorithms are tested on five real datasets ranging from 7000 to 10,000,000 in size and a simulation dataset of 100,000,000 samples. In the experiments, the implementation can scale up to datasets with 100,000,000 samples and hundreds of features and the algorithms also significantly improved the prediction accuracy when only a very small percentage of the data is labeled. The results demonstrate that the BigLP implementation is highly scalable to big data and effective in utilizing the unlabeled data for semi-supervised learning.

\end{abstract}

\section{Introduction}

Semi-supervise learning is particularly helpful when only a few labeled data points and a large amount of unlabelled data are available for training a classifier.
The unlabelled data are utilized to capture the underlying manifold structure and clusters by smoothness assumption such that the information from the labelled data points can be propagated through the clusters along the manifold structure. Graph-based semi-supervised learning algorithms perform label propagation in a positively-weighted similarity graph between the data points  \cite{Zhu03, Belkin03}. With the initialization of the vertices of the labeled data, the labels are iteratively propagated between the neighboring vertices and the propagation process will finally converge to the unique global optimum minimizing a quadratic criterion \cite{Zhou04}. To construct the similarity graph for label propagation, the commonly used and well accepted measure is Gaussian kernel similarity. The Gaussian kernel applies a non-linear mapping of the data points from the original feature space to a new infinite-dimensional space and computes a positive kernel value for each pair of data points as the similarity in the graph. Since computing the pairwise similarity between the training data is prohibitively expensive under the presence of a huge amount of unlabelled data, no general label propagation method/tool is available for learning with tens of millions or more data points.

In this paper, we propose to improve the scalability of label propagation algorithms with a method based on both low-rank approximation of the kernel matrix, and parallelization of the approximation algorithms and label propagation, named BigLP (Big-data label propagation). We first adopted two low-rank label propagation algorithms, GLNP (Global Linear Neighborhood Propagation)  \cite{tian2012global}  and Kernel Nystr\"{o}m Approximation \cite{williams2001}, and implemented the parallelized algorithms. Specifically, GLNP was accelerated with Nesterov's accelerated projected gradient descent and implemented with OpenMP for shared memory, and Kernel Nystr\"{o}m Approximation was implemented with Message Passing Interface (MPI) for distributed memory. The low-rank approximation and the parallelization of the algorithms allowed the scalability of label propagation up to 100 million samples in our experiments. The low-rank approximation of the kernel graph preserved the useful information in the original uncomputable similarity graph such that the classification results are similar or often better than the original label propagation or supervised learning algorithms that only use labeled data points. Overall, our results suggest that BigLP is effective and ready-to use implementation that will be greatly helpful for big data analysis with semi-supervise learning.

\section{Graph-based Semi-Supervised Learning}
In this section, we first review the graph-based semi-supervised learning for label propagation and then introduce the two methods for low-rank approximation of the similarity graph matrix for scalable label propagation.

\subsection{Label Propagation}\label{LP}
In a given dataset $\mathcal{X}=\{x_1, \ldots, x_l, \ldots, x_n\}$ and a given label set $\mathcal{L}=\{+1,-1\}$, $\{x_1, \ldots, x_l\}$ are data points in $\mathbb{R}^m$ labeled by $\{y_1, \ldots, y_l | y_i \in \mathcal{L}, i=1,\dots,l \}$ and $\{x_{l+1}, \ldots, x_n\}$ are unlabeled data points in $\mathbb{R}^m$. In graph-based semi-supervised learning, a similarity graph $G=(V,E)$ is first constructed from the dataset $\mathcal{X}$, where the vertex set $V=\mathcal{X}$ and the edges $E$ are weighted by adjacency matrix $W$ computed by Gaussian kernel as $W_{ij}=\exp(-\frac{\left\| x_i - x_j \right\|^2}{2\sigma^2})$, where $\sigma$ is the width parameter of the Gaussian function. Let $S = D^{-1/2}WD^{-1/2}$, where $D$ is a diagonal matrix with $D_{ii}$ equal to the sum of the $i$th row of W. By relaxing the class label variables as real numbers, label propagation algorithm iteratively updates the predicted label $f$ by
\begin{equation}\label{eqn:LP1}
	f^{t+1} = \alpha S f^t + (1-\alpha)f_0,
	\end{equation}
where $t$ is the step, and $\alpha \in (0,1)$. $f_0$ is a vector encoding the labeling of data points from set $\mathcal{L}$ and 0 is assigned to the unlabeled data. 
After running label propagation, the labels of the data points $\{x_{l+1}, \ldots, x_n\}$ are assigned based on $f^*$.


\subsection{Low-rank Label Propagation}
In large-scale semi-supervised learning, the number of samples can be in the order of tens of millions or more, leading to the difficulty in storing and operating the adjacency matrix $W$. A general solution is to generate a low-rank approximation of $W$. Specifically, the $n{\times}n$ symmetric positive semi-definite kernel matrix $W$ can be approximated by $W \approx FF^T$, where  $F \in \mathbb{R}^{n{\times}k}$ and $k {\ll} n$. Let $\bar{F}$ denotes the normalized $F$ with
\begin{equation}\label{eq:lrlp_norm}
\bar{F}_{ij}=\frac{F_{ij}}{\sqrt{F_{i,:}\sum{F}}}
\end{equation}
where $F_{i,:}$ represents row $i$ of $F$, $\sum{F}$ is a vector composed by the sum of each column of $F$, and $S \approx \bar{F}\bar{F}^T$.
With the approximation, Eqn. \eqref{eqn:LP1} can be rewritten as
\begin{equation}\label{eqn:LP2}
	f^{t+1} = \alpha{\bar{F}}{\bar{F}^{T}}f^t + (1-\alpha)f_0.
\end{equation}
In this new formula, the computational and memory requirements associated with handing the matrix $\bar{F}$ is $O(kn)$, which is much lower than $O(n^2)$. Nystr\"{o}m Method \cite{williams2001} and Global Linear Neighborhood Propagation (GLNP) \cite{tian2012global} were previously proposed to learn the low rank approximation for label propagation.

As shown in \cite{Zhou04}, the closed-form solution of Eqn. \eqref{eqn:LP2} can be directly derived
\begin{equation}\label{eqn:LP3}
f^* = \lim_{t \to \infty} f^t = (1-\alpha)(I_n-\alpha \bar{F} \bar{F}^T)^{-1}f_0,
\end{equation}
where $I_n$ denotes the $n \times n$ identity matrix. Taking advantage of the low-rank structure of $I_n-\alpha \bar{F} \bar{F}^T$,  applying Matrix-Inversion Lemma \cite{woodbury1950inverting} generates a simplified solution as
\begin{equation}\label{eqn:LP4}
(I_n-\alpha \bar{F} \bar{F}^T)^{-1}=I_n - \bar{F}(\bar{F}^T\bar{F} - (1/\alpha) I_k)^{-1}\bar{F}^T.
\end{equation}
In this solution, the $k \times k$ matrix $\bar{F}^T\bar{F} - (1/\alpha) I_k$ needs to be inverted instead of the $n \times n$ matrix $I_n-\alpha \bar{F} \bar{F}^T$. Overall, the time complexity of computing the closed-form solution $f^*$ is $O(k^3+nk)$, which is a better choice for small $k$, compared with the time complexity of iterative Eqn. \eqref{eqn:LP2} which is $O(knT)$ where $T$ is the total number of iterations for convergence.


\subsection{Nystr\"{o}m Method}

Let $W_{ij} = w(x_i,x_j)$ for a kernel function $w(a,b) = \langle\Phi{(a)},\Phi{(b)}\rangle$, where $a,b\in\mathcal{X}$ and $\Phi$ is a mapping function. The Nystr\"{o}m method generates low-rank approximations of $W$ using a subset of the samples in $\mathcal{X}$ \cite{williams2001}. Suppose $k \ll n$ data points \{ $\bar{x}_1, \bar{x}_2,...,\bar{x}_k$ \} are sampled from $\mathcal{X}$ without replacement and let $G$ be the $k{\times}k$  kernel matrix of the random samples, where $G_{i,j}=w(\bar{x}_i,\bar{x}_j)$.
Let $C$ be the $n$ by $k$ kernel matrix between $\mathcal{X}$ and the random samples, where $C_{i,g}=w(x_i,\bar{x}_g)$. The kernel matrices $W$ and $C$ can be written in blocks as
\[
W=
  \begin{bmatrix}
    G & W_{21}^{T} \\
    W_{21} & W_{22}
  \end{bmatrix} \text{and } C =  \begin{bmatrix}
    G \\
    W_{21}
  \end{bmatrix}.
\]
$G$ and $C$ can be applied to construct a rank-$k$ approximation  to $W$:
\begin{equation}\label{eqn:Nys}
	W \approx CG_k^{+}C^T  = FF^T,
\end{equation}
where $G_k^{+}$ is the pseudo-inverse of $G$ and the low rank matrix ${F} = C\sqrt{G_k^{+}}$, where $\sqrt{G_k^{+}}$ denotes element-wise square root of $G_k^{+}$, can be computed to approximate $W$ for low-rank label propagation in Eqn. \eqref{eqn:LP1}.

Instead of selecting $k$ random data points, $k$-means clustering could be applied to construct Nystr\"{o}m low-rank approximation. The $k$ centroids obtained from the $k$-means were used as the landmark points to improve the approximation over random sampling \cite{zhang2008}.

\subsection{Global Linear Neighborhood Propagation}\label{GLNP}
Another strategy to learn the low rank representation is through global linear neighborhood \cite{tian2012global}. 
Global linear neighborhood propagation (GLNP) was proposed to preserve the global cluster structures by exploring both the direct neighbors and the indirect neighbors in \cite{tian2012global}. It is shown that global linear neighborhoods can be approximated by a low-rank factorization of an unknown similarity graph. Let $X$ be the $n \times m$ data matrix from $\mathcal{X}$ where $X_{ij}$ is the value of the data point $x_i$ at the $j$th dimension. Instead of selecting $k$ neighbors to construct the similarity graph, GLNP learns a non-negative symmetric similarity graph by solving the following optimization problem:
\begin{equation}
	\label{eqn:GL_F}
	\min \mathcal{Q}(F) = \left\| X - F F^T X \right\|^2
\end{equation}
subject to $F_{ij} \geq 0$ where $F$ is a $n \times k$ matrix. 
To solve Eqn.~\eqref{eqn:GL_F}, a multiplicative updating algorithm for nonnegative matrix factorization was proposed in \cite{tian2012global}. 
Assume that $X$ contains only nonnegative values, a nonnegative $F$ can be learned by the following multiplicative update rule:
\begin{equation}
	\label{eqn:GL_F_update}
	F_{ij} \leftarrow F_{ij} \times \sqrt{\frac{(2 X X^T F)_{ij}}{(F F^T X X^T F + X X^T F F^T F)_{ij}}},
\end{equation}
where $\times$ represents element-wise multiplication. 
After $F$ is learned, it can be used for label propagation.

\begin{figure*}[!]
\centering
\begin{tabular}{c}
{\scalebox{0.45}{\includegraphics{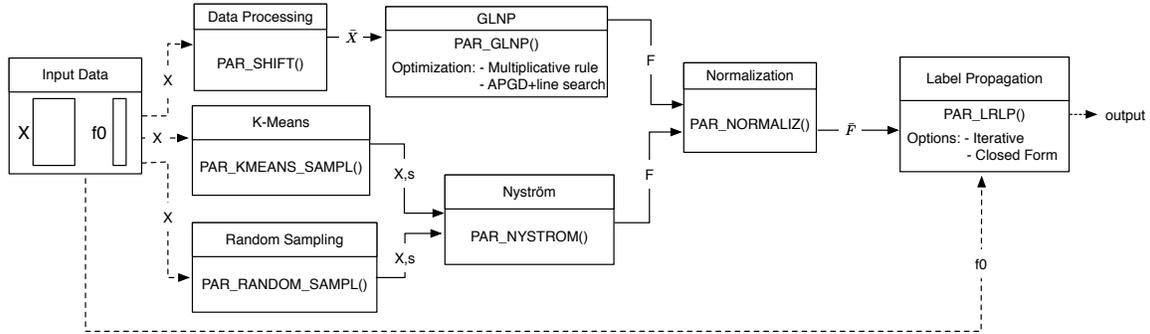}}}
\end{tabular}
\vspace{-3mm}
\caption{{\bf Diagram of the implementation architecture.} Each step is described and annotated with the function name in the parallel implementation. The input data are data matrix $X$ and initial labeling $y_0$. $X$ is first pre-processed (for GLNP). The data matrix is then used as input data for GLNP or Nystr\"{o}m algorithms using $k$-means or random sampling. Finally, the low-rank matrix is normalized and label propagation is run on the normalized low rank matrix and the input labeling $f_0$.} \label{fig:alg_diagram}
\end{figure*}

\subsection{Accelerated Projected Gradient Descent}
The objective function $\mathcal{Q}(F)$ in Eqn. \eqref{eqn:GL_F} is a fourth order non-convex function of $F$ similar to the symmetric NMF problem in \cite{conf/sdm/KuangPD12}. 
For large-scale data, a first-order optimization method is preferred to find a stationary point \cite{bertsekas1999nonlinear}.
Applying the gradient descent method $a^{r+1}=a^r-\frac{1}{L}\nabla f(a^r)$ to a convex Lipschitz continuous function $f(a)$ with $||\nabla f(a)-\nabla f(b)||\leq L||a-b||$, the rate of convergence after $r$ steps is $O(1/r)$ satisfying $f(a^r-a^*)\leq \frac{2L||a^0-a^*||^2}{r+3}$. In \cite{nesterov1983method}, an optimal first order Nesterov's method was proposed to achieve $O(1/r^2)$ convergence rate with $f(a^r-a^*)\leq \frac{2L||a^0-a^*||^2}{r^2}$.
Since Nesterov's method is often used to accelerate the projected gradient descent to solve constraint optimization problems \cite{beck2009fast, o2015adaptive}. Here we adopt Nesterov's accelerated projected gradient descent method to minimize the objective function $\mathcal{Q}(F)$ in Eqn. \eqref{eqn:GL_F} in Algorithm \ref{alg:accelerate}.

\begin{algorithm}[h]
  \begin{algorithmic}[1]
      \caption{Accelerated Projected Gradient Descent}\label{alg:accelerate}
	  \State initialize $Y^1=F^0$, $\gamma_1=1$
      \For {$t = 1 \to maxIter \do $}
        \State $F^t=P[Y^t-\alpha_t\nabla \mathcal{Q}(Y^t)/ ||\nabla \mathcal{Q}(Y^t)||]$
        \State $\gamma_{t+1}=\frac{1+\sqrt{1+4\gamma_t^2}}{2}$
        \State $Y^{t+1}=F^t+(\frac{\gamma_t-1}{\gamma_{t+1}}) (F^t-F^{t-1})$
        \If {$||\nabla^P \mathcal{Q}(F^t)|| \leq \epsilon ||\nabla \mathcal{Q}(F^0)||$}
        	\State $break$
        \EndIf
      \EndFor
      \State \textbf{return} $F$
  \end{algorithmic}
\end{algorithm}

The operation $P[C]$ denotes projecting $C$ into the nonnegative orthant such that:
\begin{equation*}
    P[C]=
    \begin{cases}
      0, & \text{if}\ \ C<0 \\
      C, & \text{otherwise}
    \end{cases}
  \end{equation*}
$\nabla^P \mathcal{Q}(F)$ is the projected gradient of variable $F$ defined as:
\begin{equation*}
    (\nabla^P \mathcal{Q}(F))_{ij}=
    \begin{cases}
      (\nabla \mathcal{Q}(F))_{ij}, & \text{if} \ \ F_{ij}\geq 0 \\
      \text{min}(0\text{, }(\nabla \mathcal{Q}(F))_{ij}), & \text{otherwise}
    \end{cases}
  \end{equation*}
The stopping condition $||\nabla^P \mathcal{Q}(F^t)|| \leq \epsilon ||\nabla \mathcal{Q}(F^0)||$ checks if a point $F^t$ is close to a stationary point in a bound-constrained optimization problem \cite{lin2007projected}.\\
The step size $\alpha_t$ in the projected gradient descent is chosen by Backtracking line search \cite{bertsekas1999nonlinear, lin2007projected} as: Given $0<\beta<1$ and $0<\sigma<1$, starting with $\alpha_1=1$ and shrinking $\alpha$ as $\alpha_{t+1}:=\beta \alpha_t$ until the condition $\mathcal{Q}(Y^{t+1})-\mathcal{Q}(Y^t) \leq \sigma \langle\nabla \mathcal{Q}(Y^t),(Y^{t+1}-Y^t)\rangle$ is satisfied.

\section{Parallel Implementation}

The architecture of the parallel implementation of the low-rank label propagation algorithms is shown in Figure \ref{fig:alg_diagram}. 
In this section, we first give a brief overview of the distributed memory and shared memory architecture, and linear algebra libraries used in the implementation, and then describe the parallel implementation of each algorithm.

\subsection{Memory Architecture}

The parallel computing approach reduces memory requirements on Label Propagation and Nystr\"{o}m low-rank matrix computation with distributed memory architecture. Shared-memory architecture was applied to run GLNP in a single computer with multi-threading.

\subsubsection{Distributed Memory:}

The distributed memory architecture follows the SPMD (single program, multiple data) paradigm for parallelism. The same program simultaneously runs on multiple CPUs according to the data decomposition. The processes communicate with each other to exchange data, as needed by the programs. The distributed memory architecture allows allocation of dedicated memory to each process possibly running on different machines for better scalability in memory requirement on each machine. The disadvantage is the overhead incurred through the data communication through the network among the machines.

Message Passing Interface (MPI) \cite{gropp1996high} was used to implement the distributed memory architecture. MPI provides a rich set of interfaces for point-to-point operations and collective communications operations (group operations). In addition, MPI-2 \cite{geist1996mpi} introduces one-sided communications operations for remote memory access. 
We used MPI to implement the parallel Low-rank Label Propagation and the Nystr\"{o}m approximation. 
In particular, the implementation of Nystr\"{o}m approximation only requires communication of size $O(n+k^2)$.

\subsubsection{Shared Memory:}

The computation of GLNP involves a large number of matrix multiplication operations which, to be performed in parallel with distributed memory, requires too much data communication.  Even if distributed memory still considerably reduces the memory requirements, the overall running time could be worse. Therefore, we adopted shared memory architecture in the implementation.

In the shared memory architecture, the program runs in multi-threading with all the threads accessing the same shared memory. There is no incurred overhead in data communication. However, the architecture can only utilize the memory available in one machine. Moreover, the shared memory architecture incurs an overhead of cache coherence, in which threads compete to access the same cache with different data, resulting in high cache misses. We implemented the shared memory architecture using the OpenMP API. 

\subsection{Linear Algebra Libraries}

In all the implementations, OpenBLAS was used to perform basic linear algebra operations. OpenBLAS is an optimized version of the BLAS library, and allows multi-threading implementation. For more advanced linear algebra operations, in the eigen-decomposition for Nystr\"{o}m Approximation, we used the LAPACK library.

\subsection{Parallel Nystr\"{o}m Approximation}

The parallel Nystr\"{o}m approximation algorithm implements both random and $k$-means sampling of $k$ samples to calculate the low-rank representation. Algorithm S.3 in the Supplementary document describes sampling $k$ random samples without replacement. Algorithm S.4 selects $k$ samples as the centroids learned by $k$-means. For improved efficiency, we typically only run $k$-means with a small number of iterations, which usually generates reasonably good selection.

\begin{algorithm}
  \begin{algorithmic}[1]
    \caption{Parallel Nystr\"{o}m}\label{alg:Nystrom}
    \Function{Par\_Nystr\"{o}m}{$X^p,X_k^p,m,n,k,maxIter$}
      \For {$i=0 \to k-1 \do$}
        \State $MPI\_Broadcast({X_k^p}_i, sample, kIdxs_i)$
        \State $W_i=RBF(X_k^p, sample)$
        \State $C_i=RBF(X^p, sample)$
      \EndFor
      \State $MPI\_Gather(W, 0)$
      \If {$rank = 0$}
      	\State $[eigvals, eigvecs] = EIG(W)$
      \EndIf
      \State $MPI\_Broadcast(eigvals, 0)$
      \State $MPI\_Broadcast(eigvecs, 0)$
      \State $G = C*eigvecs$
	  \For {$i=0 \to k-1 \do$}
	    \State $G_i = G_i / \sqrt{eigvals_i}$
	  \EndFor
      \State \textbf{return} $G$
    \EndFunction
  \end{algorithmic}
\end{algorithm}

Based on the selected $k$ samples, Nystr\"{o}m approximation algorithm is implemented in Algorithm \ref{alg:Nystrom}. In Algorithm \ref{alg:Nystrom}, the process assigned with sample $i$ broadcasts sample $i$ to the other processes (line 3). After receiving sample $i$, each process calculates $W$ and $C$ entries between sample $i$ and all the samples at the node, with RBF kernel (lines 4-5). Matrix $W$ is then gathered by process 0 to perform the eigen-decomposition of $W$ (lines 7-10). Note that since $W$ is only $k \times k$, the eigen-decomposition is not expensive for small $k$. Process 0 then broadcasts the eigenvectors and eigenvalues to the other processes at lines 11-12. Each process finally calculates the G based on the received eigenvectors and eigenvalues (lines 13-16).

\subsection{Parallel GLNP}
We implemented parallel GLNP following the two optimization frameworks presented previously: multiplicative update rule and accelerated projected gradient descent with line search. In the multiplicative update rule, given the input data matrix $X$, the function PAR\_SHIFT() in Algorithm S.1 checks the minimum value of $X$ and then adds the minimum value to $X$ to obtain the non-negative matrix $\bar{X}$ since GLNP is based on non-negative multiplicative updating. The implementation of GLNP using multiplicative update rule is described in Algorithm \ref{alg:glnp}.

\begin{algorithm}
  \begin{algorithmic}[1]
    \caption{Parallel GLNP - Multiplicative update rule}\label{alg:glnp}
    \Function{Par\_GLNP\_MUL}{$X,m,n,k,maxIter,tol$}
	  \State $F \gets U_{m \times k}[0,1]$
      \For {$t = 0 \to maxIter \do $}
        \State $F_{old} = F$
        \State $B = X(X^TF)$
        \State $D = F(F^TB)$
        \State $G = B(F^TF)$
        \For {$i = 0 \to m-1 \do $}
        	\For {$j = 0 \to k-1 \do $}
        		\State $F_{ij} = F_{ij} sqrt(2B_{ij}/(D_{ij}+G_{ij}))$
        	\EndFor
        \EndFor
        \If {$max(abs(F_{old} - F)) < tol$}
        	\State $break$
        \EndIf
      \EndFor
      \State \textbf{return} $F$
    \EndFunction
  \end{algorithmic}
\end{algorithm}

In Algorithm \ref{alg:glnp}, $F$ is first randomly initialized with uniform distribution between 0 and 1 in parallel by OpenMP. Then, the multiplicative update rule in Eqn.~\eqref{eqn:GL_F} is decomposed into several steps of matrix multiplication for parallelization according to the data dependency (lines 5-7). These operations are performed in multi-threading by the OpenBLAS library. Note that all these multiplications are computed in $O(kn)$. Lines 8-12 update $F$ with the multiplicative rule using the intermediate results in $B$, $C$ and $D$ with openMP.  Lines 13-15 check for convergence by the threshold $tol$. Instead of checking the convergence of the objective function, which increase the memory requirements, the algorithm checks the maximum change among the elements in $F$. In our observation, the convergence is always achieved with this criteria.

\begin{algorithm}
  \begin{algorithmic}[1]
    \caption{Parallel GLNP - Projected Gradient Descent with Line Search}\label{alg:glnp_pgd}
    \Function{Par\_GLNP\_APGD}{$X$,$m$,$n$,$k$,$maxIter$, ,$maxInnerIter$,$beta$,$tol$,$roll$}
	  \State $F \gets U_{m \times k}[0,1]$
	  \State $Y = F$
      \For {$t = 0 \to maxIter \do $}
        \State $B = X(X^TY)$
        \State $D = Y(Y^TB)$
        \State $G = B(Y^TY)$
        \For {$i = 0 \to m-1 \do $}
        	\For {$j = 0 \to k-1 \do $}
        		\State $Grad_{ij} = 2D_{ij} + 2G_{ij} - 4B_{ij}$
        	\EndFor
        \EndFor
        \State $Grad_0 = Grad$
        \State $Grad = Grad / sqrt(sum(Grad))$
        \State $obj_{old} = obj$
        \State $obj = ||X-Y^TYX||^2$
        \State $alpha = 1$
        \For {$inner = 0 \to maxInnerIter \do $}
        	\State $Y_1 = max(Y-alpha . Grad, 0)$
        	\State $obj_1 = ||X-Y_1^TY_1X||^2$
        	\State $sum = \sum(Grad_0 * (Y_1-Y))$
        	\If {$obj_1-obj < roll . sum$}
        		\State $break$
        	\EndIf
        	\State $alpha = beta^{inner+1}$
        \EndFor
        \State $F_{old} = F$
      	 \State $F = Y_1$
        \State $t_{old} = t$
        \State $t = (1+sqrt(1+4t^2))/2$
        \State $Y = F + (F-F_{old})(t_{old}-1)/t$
        \If {$abs((obj_1-obj_{old})/obj_1) < tol$}
        	\State $break$
        \EndIf
      \EndFor
      \State \textbf{return} $F$
    \EndFunction
  \end{algorithmic}
\end{algorithm}
\begin{table}[!]
\scriptsize
\centering
\begin{tabular}{|c|ccccc|}
\hline
{Dataset}&{HEPMASS}&{SUSY}&{mnist8m}&{Protein}&{Gisette}\\
\hline
{Sample}&{$10.5 \times 10^6$}&{$5 \times 10^6$}&{1,648,890}&{13,077}&{7,000}\\
{Feature}&{27}&{128}&{784}&{357}&{5,000}\\
\hline
\end{tabular}
\caption{Summary of datasets}\label{tab:summary}
\end{table}

The GLNP implementation with projected gradient descent and line search is presented in Algorithm \ref{alg:glnp_pgd}. In Algorithm \ref{alg:glnp_pgd}, we first calculate the normalized and unnormalized gradient of the objective function (lines 9-15). Line 16 calculates the objective function used by the line search. Lines 18-26 will perform the inner iterations of the projected gradient descent. Finally,  the convergence is checked on line 33.

\subsection{Parallel Low-rank Label Propagation}

After normalizing low rank matrix $F$ by the function PAR\_NORMALIZ() in Algorithm S.2 according to Eqn. \eqref{eq:lrlp_norm}, parallel low-rank label propagation is performed on the normalized low-rank data $\bar{F}$ and the initial labeling vector $f^{0} \in \mathbb{R}^{n \times 1}$ with Algorithm S.5. Note that $f^0$ is also divided among the processes such that each process contains only a vector ${f^{0}}^p \in \mathbb{R}^{\frac{n}{p} \times 1}$. Algorithm S.5 first initializes $f^p$ by sampling an uniform distribution between -1 and 1 (line 2). Each process is only responsible for calculating the allocated part of $f$. Lines 5-7 perform label propagation, and lines 8-12 check for convergence. Each process will return the local $f^p$.

\section{Results}

The parallel algorithms are tested on five real datasets and a simulation dataset. The runtime and memory requirement are measured. The prediction accuracy for semi-supervised learning was also reported.

\subsection{Datasets}
Five datasets with various sample sizes and feature sizes described in Table \ref{tab:summary} were downloaded. The two largest datasets, HEPMASS and SUSY, were downloaded from UCI. Each of them contains millions of samples but a small number of features. mnist8m is the handwritten digit data from \cite{loosli2007} which contains digits 7 and 9 for classification. The Protein dataset is for protein secondary structure prediction. In the experiments we only selected two out of the three classes for classification. The Gisette dataset is also a handwritten digit dataset used for feature selection challenge in NIPS 2003. Finally, we also created a random simulation dataset, with 100 million samples and 100 features to test the scalability of the implementation.

\subsection{Runtime and Memory Requirements}
\begin{figure*}[h!]
\centering
\begin{tabular}{c}
{\scalebox{0.5}{\includegraphics{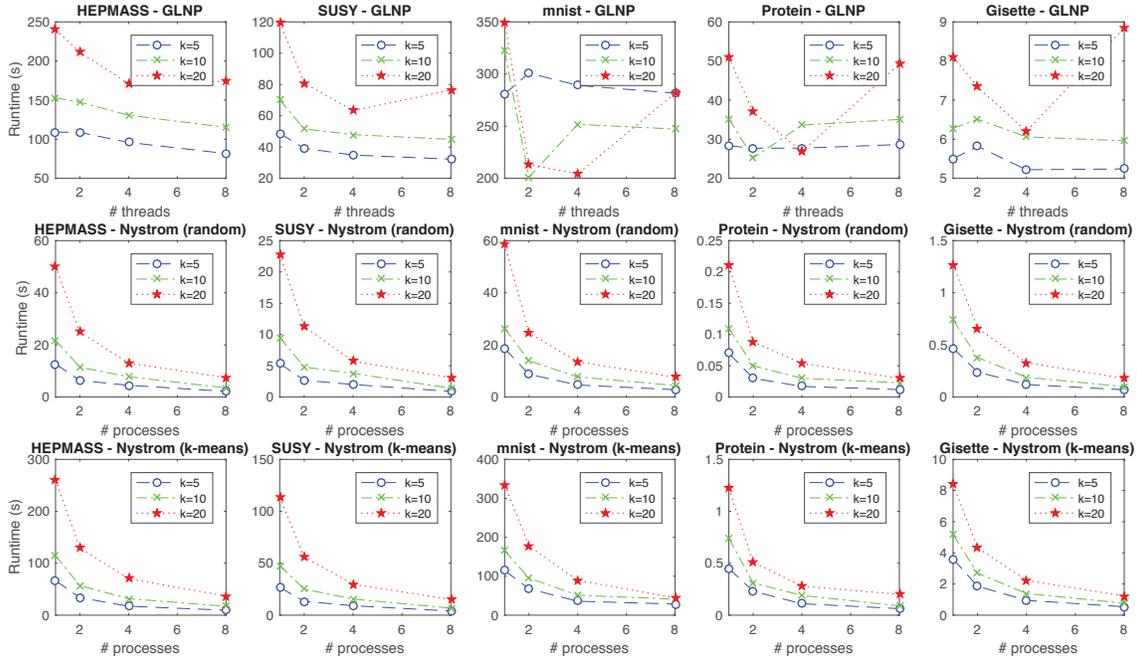}}}
\end{tabular}
\vspace{-3mm}
\caption{{\bf Runtime of GLNP, Nystr\"{o}m (Random) and Nystr\"{o}m ($k$-means).}} \label{fig:runtime}
\end{figure*}

\begin{figure*}[h!]
\centering
\begin{tabular}{c}
{\scalebox{0.5}{\includegraphics{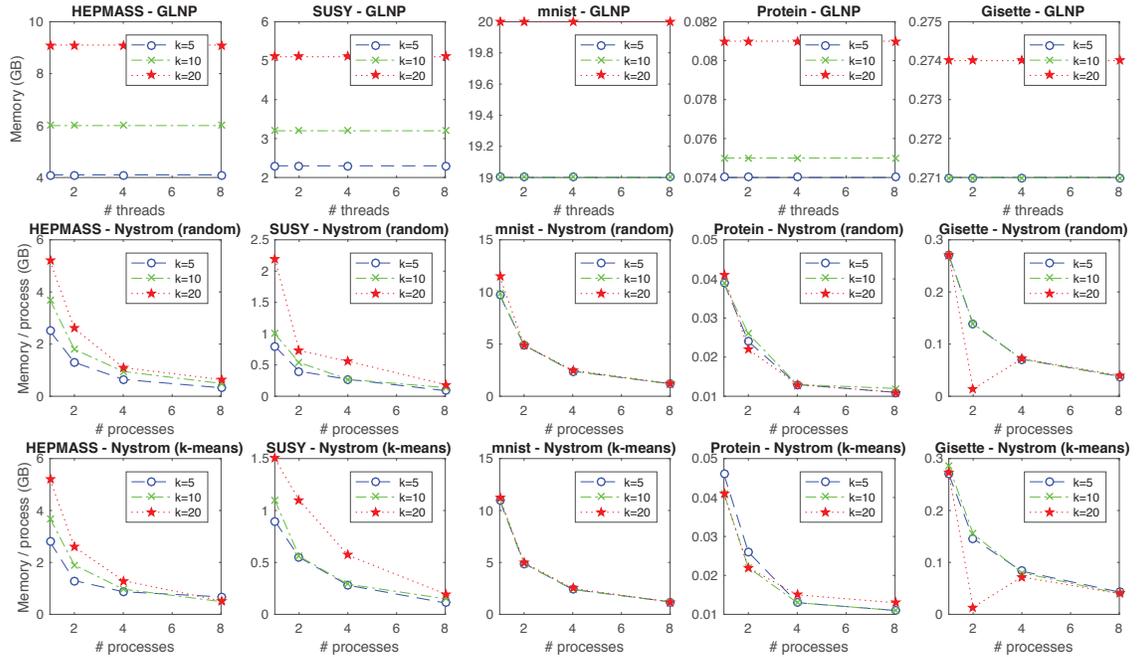}}}
\end{tabular}
\vspace{-3mm}
\caption{{\bf Memory requirements of GLNP, Nystr\"{o}m (Random) and Nystr\"{o}m ($k$-means).}} \label{fig:memory}
\end{figure*}

We measured the runtime and memory requirements of our parallel implementation of Nystr\"{o}m (both random sampling and $k$-means sampling) and GLNP in all the datasets, shown in Figures \ref{fig:runtime} and \ref{fig:memory}. 

Figure \ref{fig:runtime} shows that GLNP is more scalable up to 4 threads and becomes worst at 8 threads due to the overhead by cache coherence with different threads competing to access the same cache which results in many cache misses. In the SUSY dataset, parallel GLNP with $k=20$ runs 1.89x faster than the serial implementation. In the HEPMASS with $10.5$ millions samples, parallel GLNP is 1.71x faster than the serial implementation. The multithreading by 4 threads clearly reduces the runtime considerably. GLNP was implemented in the shared-memory architecture, which always requires a constant amount of memory independent of the number of threads in Figure \ref{fig:memory}.

Figure \ref{fig:runtime} also confirms that Nystr\"{o}m is a very scalable algorithm. Using 8 processes, the parallel implementation of the random sample selection with $k=20$ performs 7.67x faster than the serial implementation on the mnist8m dataset, and 7.48x faster with sample selection by $k$-means. In the HEPMASS dataset, the algorithm was 7.08x faster using random sampling, and 7.42x using $k$-means. In Figure \ref{fig:memory}, the Nystr\"{o}m implementation reduces the memory requirements on each machine with the distributed memory architecture without introducing much overhead consumption. Note that among the large datasets, mnist dataset has relative more features. The memory consumption for different $k$ is very similar since the original dataset is larger than the low-rank approximation data by a big magnitude.

\begin{figure*}[!tpb]
\begin{tabular}{ccc}
{\scalebox{0.26}{\includegraphics{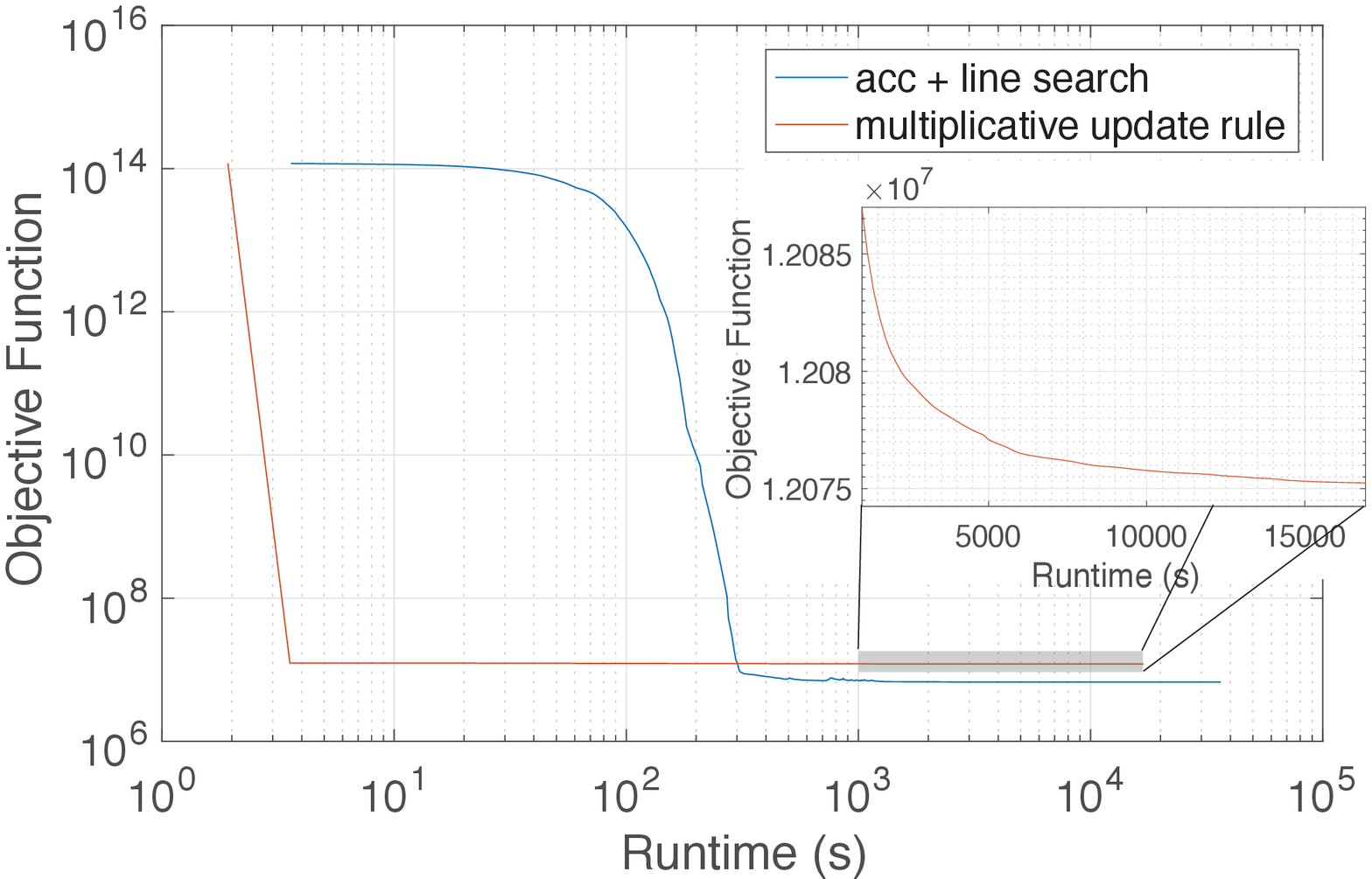}}} & {\scalebox{0.26}{\includegraphics{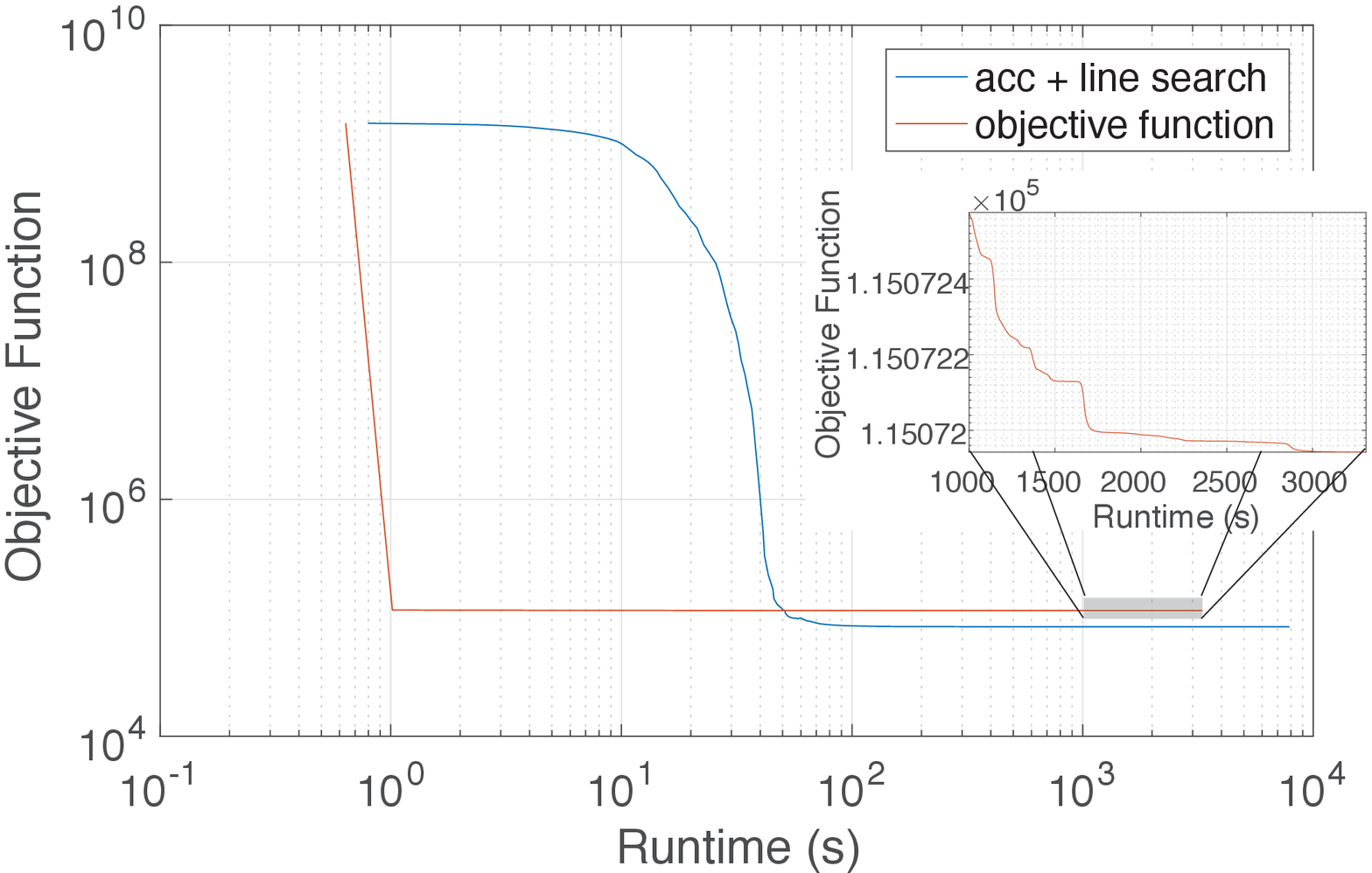}}} & {\scalebox{0.26}{\includegraphics{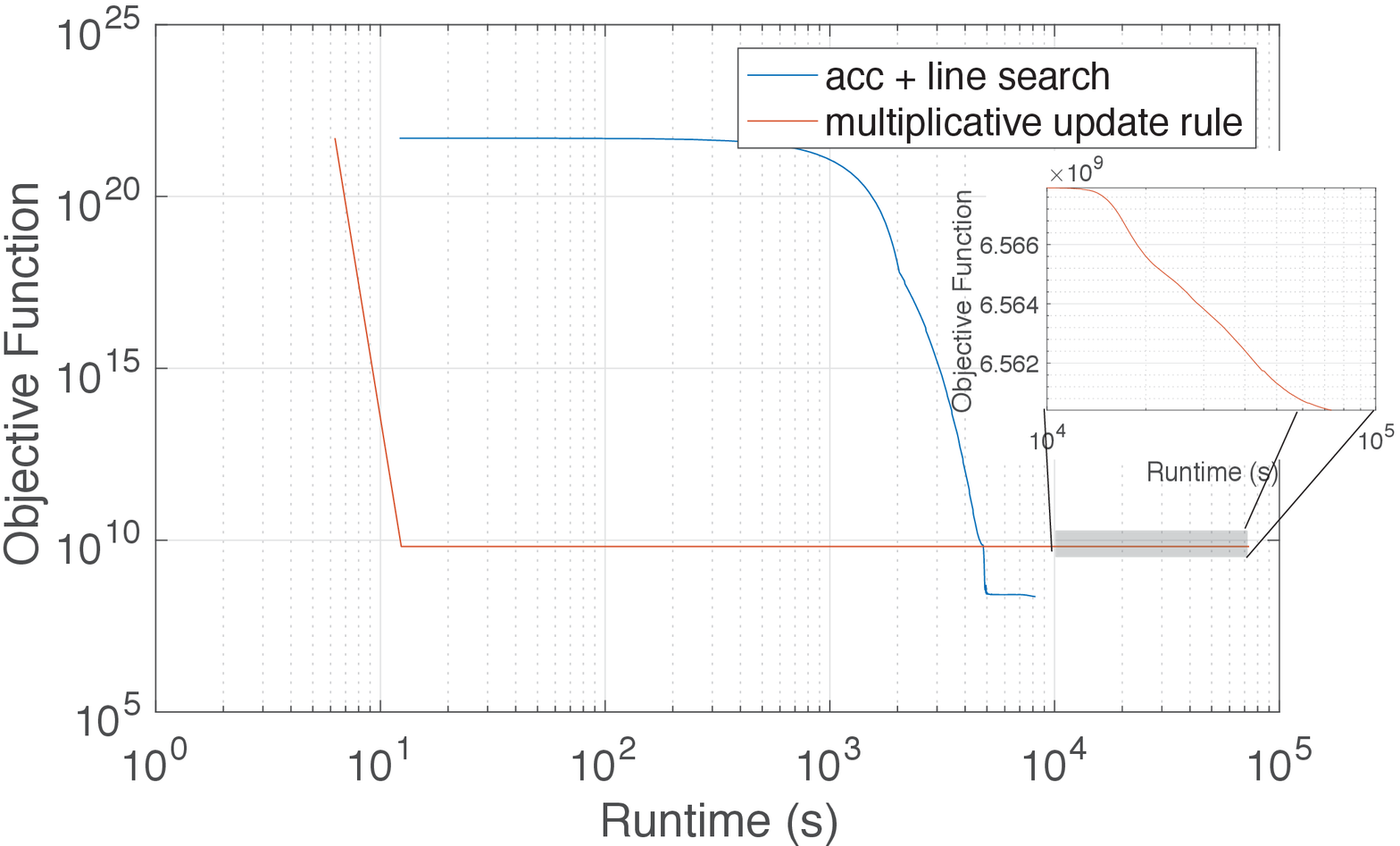}}}\\
(A) Gisette k=100 &
(B) Protein k=100 &
(C) HEPMASS k=10 \\
\end{tabular}
\caption{{\bf Comparison of optimization techniques for GLNP.} The plots show a comparison of optimization by accelerated projected gradient descent with linear search and multiplicative updating on three datasets. 
}
\label{fig:obj_funcs}
\end{figure*}

In Figure \ref{fig:obj_funcs}, the plots show a comparison of the optimization by GLNP with acceleration plus line-search and multiplicative updating on three datasets Gisette, Protein and HEPMASS. In all the three cases, accelerated projected gradient descent achieved a better local optimal. Multiplicative updating has a very fast drop in the objective function in the first iteration and then gets into very slow steps for convergence. In practice, we observed that accelerated projected gradient descent achieves better local optimal and convergence in less iterations in all the experiments.

Finally, we evaluated the performance on the simulation dataset with 100 millions of samples and 100 features. We were able to run this dataset using at least 8 processes by the Nystr\"{o}m implementation. With $k$=20 under random sample selection, the implementation completes in 140 seconds with 8 processes. The implementation under $k$-means sample selection runs in 543 seconds with 16 processes. It is also important to note that the memory requirements by each process is only 6.5 GB when 16 processes are used, which allows the implementation to run even on most personal computers available nowadays.

\begin{figure*}[h!]
\centering
\begin{tabular}{c}
{\scalebox{0.165}{\includegraphics*{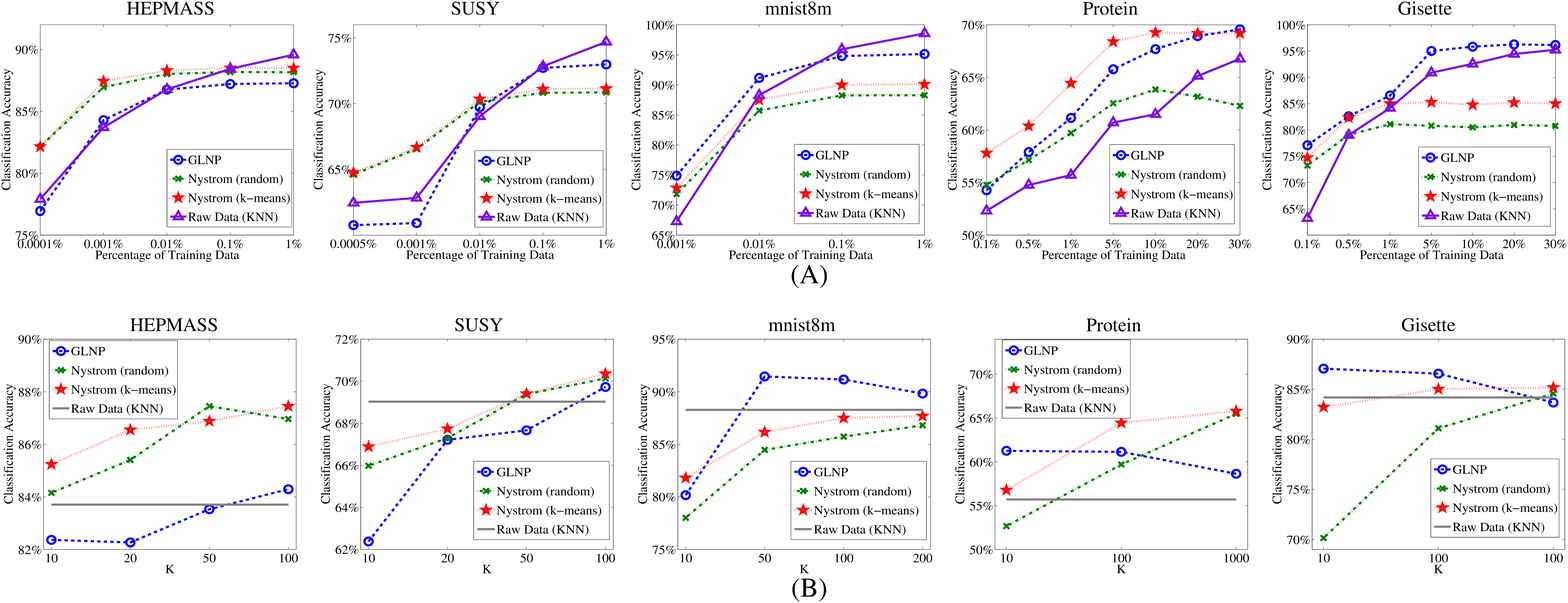}}}
\end{tabular}
\vspace{-3mm}
\caption{{\bf Classification results on five datasets.} (A) The x-axis shows percentage of training samples. (B) The x-axis shows $k$ selected to approximate the low-rank matrices.} \label{fig:classification}
\end{figure*}

\subsection{Classification on Five Datasets}
To test the performance of semi-supervised learning with low-rank matrix approximation, we compared label propagation on the low-rank matrices approximated by GLNP and Nystr\"{o}m approximation (both random sampling and $k$-means sampling) with the $k$-nearest neighbor (KNN) classification algorithm on the original data by considering the five nearest training samples. 
To evaluate the classification results,  we tested different $k$ for low-rank approximation.
 In the experiments, we held out 20\% of samples as the test set, and randomly selected different percentages of samples as the training set in each trail. On each dataset, for each $k$ and each percentage of training samples, we ran 10 trails with different randomly selected training data and report the average classification accuracy on the test set. The same setup was applied to test KNN as a base line. In label propagation, $\alpha$ was set to 0.01.

The classification results are reported in Figure \ref{fig:classification}. In Figure \ref{fig:classification}(A), $k$ was fixed to 100 for each experiment and the plots show the results of training with different percentages of training samples. In general, semi-supervised learning by label propagation with low-rank matrix approximation performs better than KNN when only a small size of training data is available. As the size of training data increases, KNN based on all the original features can perform similarly or better on the large datasets. The observation is consistent with the assumption of semi-supervised learning that the underlying manifold structure among labeled and unlabeled data can be explored to improve classification of unlabeled samples when only a small amount of training data is available. As more and more samples become available for training, the structural information becomes less important. Furthermore, low-rank matrix approximation can potentially lose information in the original dataset when $k$ is small. Thus, it is possible that the classification results with low-rank label propagation could be slightly worse than KNN when the size of training data is large. Another observation is that the performance of GLNP is better than Nystr\"{o}m on the small datasets but worse on the large ones. It is possibly because GLNP often requires more iterations to learn the low-rank matrix and convergence is more difficult to achieve on the large datasets. Finally, consistent with previous observations, Nystr\"{o}m with $k$-means sampling consistently is better than random sampling.

In Figure \ref{fig:classification}(B), the number of training samples were fixed to around 100 for each dataset and results show the effect of choosing different rank $k$. In general, as the size of $k$ increase, the classification performances of low-rank approximation algorithms are closer to the baseline method. In addition, as $k$ increases, the classification performances of Nystr\"{o}m, both $k$-means and random sampling, become better. It is also noticeable that the performance of GLNP is less sensitive to the parameter $k$ since it relies on the global information. Overall, the classification performances of low-rank label propagation are very competitive or better than supervised learning algorithm KNN using the original feature space when $k$ is sufficient. Furthermore, for the largest three datasets, KNN is only scalable to use up to 1$\%$ of samples as training data while the low-rank label propagation are scalable to use all of the training data.

\section{Discussion}
In this paper, we applied low-rank matrix approximation and Nesterov's accelerated projected gradient descent with parallel implementation for Big-data Label Propagation (BigLP). BigLP was implemented and tested on the datasets of huge sample sizes for semi-supervised learning. Compared with sparsity induced measures \cite{cheng2009} to construct similarity graphs, BigLP is more applicable to the datasets of huge sample size with a relatively small number of features that need to be kernelized for better classification in semi-supervised learning. Sparsity induced measures rely on knowing all the pairwise similarities and would not scale to the datasets with more than hundreds of thousands of samples due to the low scalability in sample size and optimization for sparsity. In addition, compared with the sparsity induced measures and local linear embedding method \cite{roweis2000}, in which the neighbors are selected ``locally", GLNP preserves the global structures among the data points, and construct more robust and reliable similarity graphs for graph-based semi-supervised learning. In terms of scalability of the two low-rank approximation methods, Nystr\"{o}m approximation is potentially better than GLNP depending on the iterations of $k$-means for sample selection. In practice, the quality of the similarity matrix constructed by Nystr\"{o}m method could also depend on the samples learned by $k$-means which could introduce uncertainty.

\section{Funding}
The research work is supported by grant from the National Science Foundation (IIS 1149697). RP is also supported by CAPES Foundation, Ministry of Education of Brazil (BEX 13250/13-2).

\bibliographystyle{siam}
\bibliography{document}{}
\end{document}


%






\section{Algorithms}

\subsection{Data Processing}

In the first step in the parallel low-rank label propagation, each feature is individually shifted to contain only non-negative numbers. Given a low-rank matrix $X \in \mathbb{R}^{n \times k}$, each process $p$ will contain a chunk of $X$, $X^p \in \mathbb{R}^{\frac{n}{p} \times k}$. Each process performs shift of its data. Algorithm S.1 shows that each process first calculates the minimum value for each feature in line 3, followed by a MPI \textit{All Reduce} operation (line 4) to give each process the minimum value among all the processes, for each feature. If the number obtained is negative, that feature is then shifted to avoid the negative number (lines 5-7). The algorithm is outlined below:


\begin{algorithm}
  \begin{algorithmic}[1]
    \caption{Parallel data shift for label propagation}\label{alg:lrlp_shift}
    \Procedure{Par\_Shift}{$X^p,n,k$}
      \For {$i = 0 \to k-1 \do $}
        \State $colMin = \min_j{x^p_{ji}}$ \Comment{minimum of column i}
        \State $MPI\_Allreduce(colMin, 1, MPI\_MIN)$
        \If {$colMin < 0$}
        	\State $x^p_{i} = x^p_{i} - colMin $
        \EndIf
      \EndFor\label{euclidendwhile}
      \State \textbf{return} $X^p$
    \EndProcedure
  \end{algorithmic}
\end{algorithm}

\subsection{Data Normalization}

The normalization was implemented in MPI, also following the assumption that each process p contains only a chunk $X^p$ of X. Algorithm S.2 outlines the implementation.
In algorithm S.2, we first build a vector containing the sum for each column (lines 2-4) followed by a MPI \textit{All Reduce} operation (line 5), which updates the vector with the sum of the vectors in all the processes. Lines 6-9 apply the normalization equation in the data in each process. The data normalization algorithm is outlined bellow:

\begin{algorithm}
  \begin{algorithmic}[1]
    \caption{Parallel data normalization for label propagation}\label{alg:lrlp_norm}
    \Procedure{Par\_Normalization}{$X^p,n,k$}
      \For {$i = 0 \to k-1 \do $}
        \State $colSum_i = \sum_jx^p_{ji}$ \Comment{sum of column i}
      \EndFor
      \State $MPI\_Allreduce(colSum, k, MPI\_SUM)$
      \State $tmp = X^p * colSum$
      \For {$i = 0 \to \frac{n}{p}-1 \do $}
        \State $x^p_i = x^p_i / \sqrt{tmp_i} $
      \EndFor\label{euclidendwhile}
      \State \textbf{return} $X^p$
    \EndProcedure
  \end{algorithmic}
\end{algorithm}
\newpage

\vspace{-15pt}

\subsection{Random sampling}
In lines 2-6, process 0 first select $k$ indices. This list is then sent to all processes (line 7). After, each process will look up and return the subset of the k indices which refers to data present in that process (lines 8-13). The algorithm is outlined below:
\begin{algorithm}
  \begin{algorithmic}[1]
    \caption{Parallel Random Sampling}\label{alg:Nystrom_random}
    \Function{Par\_Random\_Sampl}{$X^p,m,n,k$}
      \If {$rank = 0$} \Comment{if process 0}
	    \State $tmp = [0 \dots (m-1)]$
	    \State $tmp = shuffle(tmp)$
		\State $kIndices = tmp[0 \dots (k-1)]$
	  \EndIf
	  \State $MPI\_Broadcast(kIndices,0)$
      \For {$i = 0 \to k-1 \do $}
      	\If {$kIndices_i/(m/p) = rank$}
  	      \State $X^p_k.insert(X^p[kIndices_i\%(m/p)])$
  	    \EndIf
      \EndFor
      \State \textbf{return} $[X^p_k, kIndices]$
    \EndFunction
  \end{algorithmic}
\end{algorithm}

\vspace{-15pt}
\subsection{$k$-means sampling}
The algorithm is divided into two sections. First, each process assigns the closest centroid to its local data points (lines 6-17). For each centroid $i$, the process containing it broadcasts the centroid to other processes (line 12). Each process then calculated the distance of its data points to the centroid received (line 13-15). In the next part, the algorithm finds new centroids based on the datapoints assignment (lines 20-41). For each centroid i, each process finds how many local datapoints belongs to it. The total number of datapoints in the centroid is then obtained by a MPI \textit{All reduce} operation (line 29). Then, the mean of all the datapoints is obtained (lines 30-26), resulting in the new centroid $i$.

\begin{algorithm}
  \begin{algorithmic}[1]
    \caption{Parallel $k$-means Sampling}\label{alg:Nystrom_kmeans}
    \Function{Par\_Kmeans\_Sampl}{$X^p,m,n,k,maxIter$}
      \State $[centrs, kIdxs] = Par\_Kmeans\_Sampling(X^p,m,n,k)$ \Comment {find random centroids}
      \For {$t=0 \to maxIter \do$}  \Comment{assign centroid to datapoints}      
        \For {$i=0 \to k-1 \do $}
          \If {$kIdxs_i / (m/p) = rank$}
		    \State $centr = centrs.next()$
          \EndIf
          \State $MPI\_Broadcast(centr, kIdxs_i / (m/p))$
          \State $dist = ||centr - X^p||^2$
        \EndFor
        \State $minIdxs = min(dist)$
        \For {$i=0 \to k-1 \do $} \Comment {find new k centroids}
          \For {$j=0 \to (m/p)-1 \do$}
            \If {$minIdxs_j = i$}
            	\State $samplesIdx.insert(j)$
            \EndIf
          \EndFor
          \State $nSamples = samplesIdx.length$
          \State $MPI\_Allreduce(nSamples, $
          \State $nTotalSamples, MPI\_SUM)$
          \For {$j=0 \to nSamples-1 \do$}
            \For {$l=0 \to n-1 \do$}
              \State $s = samplesIdx_j$
              \State $newCtr_l = X^p_{sl} / nTotalSamples$
            \EndFor
          \EndFor
          \State $MPI\_Reduce(newCtr, kIdxs_i / (m/p), $
          \State $MPI\_SUM)$
          \If {$kIdxs_i/(m/p)$}
            \State $centrs.replace(newCtr)$
          \EndIf
        \EndFor
      \EndFor
      \State \textbf{return} $centrs$
    \EndFunction
  \end{algorithmic}
\end{algorithm}

\newpage

\subsection{Parallel low-rank label propagation}

The parallel low-rank label propagation algorithm is shown below. In the algorithm, $X$ and $f^0$ are divided among the processes such that each process contains only a matrix ${X}^p \in \mathbb{R}^{\frac{n}{p} \times n}$ and a vector ${f^{0}}^p \in \mathbb{R}^{\frac{n}{p} \times 1}$. The algorithm first initializes $f^p$ of size $\frac{n}{p}$ with an uniform distribution between -1 and 1 (line 2). Each process is only responsible for calculating the allocated part of $f$. Lines 5-7 perform label propagation, and lines 8-12 check for convergence. Each process will return the local vector $f^p$.

\begin{algorithm}
  \begin{algorithmic}[1]
    \caption{Parallel low-rank label propagation}\label{alg:lrlp}
    \Function{Par\_LRLP}{$X^p,{f^{0}}^p,n,k,\alpha,maxIter,tol$}
	  \State $f^p \gets U_{\frac{n}{p}}[-1,1]$
      \For {$t = 0 \to maxIter-1 \do $}
        \State $f_{old}^p = f^p$
        \State $f_{tmp}^p = (X^p)^T*f^p$
        \State $MPI\_Allreduce(f_{tmp}^p, k, MPI\_SUM)$
        \State $f = \alpha*X^p*f_{tmp}^p + (1-\alpha)*{f^{0}}^p$
        \State $tmp = max(abs(f_{old}^p - f^p))$
        \State $MPI\_Allreduce(tmp, 1, MPI\_MAX)$
        \If {$tmp < tol$}
        	\State $break$
        \EndIf
      \EndFor
      \State \textbf{return} $f^p$
    \EndFunction
  \end{algorithmic}
\end{algorithm}

\newpage

\section{Classification Results}
The classification results tested under different choice of $k$ and percentage of training samples are shown in Tables \ref{tab:HEPMASS}-\ref{tab:Gisette} for the five datasets, respectively.

\begin{table*}[!]
\scriptsize
\centering
\begin{tabular}{|c|cccccc|cc|}
\hline
{}&\multicolumn{6}{c|}{Low rank approximation}&\multicolumn{2}{c|}{Raw Data}\\
{}&\multicolumn{2}{c}{GLNP}&\multicolumn{2}{c}{Nystr\"{o}m (random)}&\multicolumn{2}{c|}{Nystr\"{o}m (k-means)}&\multirow{ 2}{*}{KNN}&\multirow{2}{*}{LP (linear)}\\
{\% of training}&{k=10}&{k=100}&{k=10}&{k=100}&{k=10}&{k=100}&&\\
\hline
{0.0001\%}&{82.71\%}&{76.95\%}&{79.22\%}&{82.16\%}&{83.49\%}&{82.15\%}&{77.91\%}&{80.62\%}\\
{0.001\%}&{82.37\%}&{84.29\%}&{84.15\%}&{86.96\%}&{85.24\%}&{87.45\%}&{83.71\%}&{85.75\%}\\
{0.01\%}&{82.37\%}&{86.76\%}&{85.11\%}&{88.03\%}&{86.10\%}&{88.30\%}&{86.78\%}&{87.06\%}\\
{0.1\%}&{82.92\%}&{87.21\%}&{85.22\%}&{88.18\%}&{86.18\%}&{88.51\%}&{88.44\%}&{87.33\%}\\
\hline
\end{tabular}
\caption{HEPMASS}\label{tab:HEPMASS}
\end{table*}

\begin{table*}[!]
\scriptsize
\centering
\begin{tabular}{|c|cccccc|cc|}
\hline
{}&\multicolumn{6}{c|}{Low rank approximation}&\multicolumn{2}{c|}{Raw Data}\\
{}&\multicolumn{2}{c}{GLNP}&\multicolumn{2}{c}{Nystr\"{o}m (random)}&\multicolumn{2}{c|}{Nystr\"{o}m (k-means)}&\multirow{ 2}{*}{KNN}&\multirow{2}{*}{LP (linear)}\\
{\% of training}&{k=10}&{k=100}&{k=10}&{k=100}&{k=10}&{k=100}&&\\
\hline
{0.0005\%}&{57.68\%}&{60.76\%}&{61.01\%}&{64.61\%}&{62.43\%}&{64.77\%}&{62.46\%}&{65.39\%}\\
{0.001\%}&{57.58\%}&{60.92\%}&{63.75\%}&{66.56\%}&{64.36\%}&{66.68\%}&{62.83\%}&{66.99\%}\\
{0.01\%}&{62.38\%}&{69.73\%}&{65.98\%}&{70.13\%}&{66.89\%}&{70.36\%}&{69.03\%}&{69.03\%}\\
{0.1\%}&{63.63\%}&{72.72\%}&{66.25\%}&{70.83\%}&{67.42\%}&{71.11\%}&{72.84\%}&{69.55\%}\\
\hline
\end{tabular}
\caption{SUSY}\label{tab:SUSY}
\end{table*}

\begin{table*}[!]
\scriptsize
\centering
\begin{tabular}{|c|cccccc|cc|}
\hline
{}&\multicolumn{6}{c|}{Low rank approximation}&\multicolumn{2}{c|}{Raw Data}\\
{}&\multicolumn{2}{c}{GLNP}&\multicolumn{2}{c}{Nystr\"{o}m (random)}&\multicolumn{2}{c|}{Nystr\"{o}m (k-means)}&\multirow{ 2}{*}{KNN}&\multirow{2}{*}{LP (linear)}\\
{\% of training}&{k=10}&{k=100}&{k=10}&{k=100}&{k=10}&{k=100}&&\\
\hline
{0.001\%}&{73.11\%}&{74.90\%}&{66.65\%}&{71.85\%}&{69.31\%}&{72.86\%}&{67.28\%}&{71.96\%}\\
{0.01\%}&{80.17\%}&{91.16\%}&{78.015\%}&{85.73\%}&{81.81\%}&{87.51\%}&{88.28\%}&{85.44\%}\\
{0.1\%}&{80.45\%}&{94.80\%}&{79.66\%}&{88.25\%}&{83.50\%}&{89.99\%}&{95.93\%}&{87.67\%}\\
{1\%}&{80.54\%}&{95.13\%}&{80.013\%}&{88.28\%}&{83.31\%}&{90.10\%}&{98.56\%}&{87.74\%}\\
\hline
\end{tabular}
\caption{mnist8m}\label{tab:mnist8m}
\end{table*}

\begin{table*}[!]
\scriptsize
\centering
\begin{tabular}{|c|cccccc|cc|}
\hline
{}&\multicolumn{6}{c|}{Low rank approximation}&\multicolumn{2}{c|}{Raw Data}\\
{}&\multicolumn{2}{c}{GLNP}&\multicolumn{2}{c}{Nystr\"{o}m (random)}&\multicolumn{2}{c|}{Nystr\"{o}m (k-means)}&\multirow{ 2}{*}{KNN}&\multirow{2}{*}{LP (RBF)}\\
{\% of training}&{k=10}&{k=100}&{k=10}&{k=100}&{k=10}&{k=100}&&\\
\hline
{0.1\%}&{56.61\%}&{54.27\%}&{51.51\%}&{54.78\%}&{54.31\%}&{57.80\%}&{52.31\%}&{57.94\%}\\
{0.5\%}&{57.81\%}&{57.90\%}&{51.32\%}&{57.14\%}&{55.02\%}&{60.38\%}&{54.77\%}&{64.15\%}\\
{1\%}&{61.26\%}&{61.14\%}&{52.67\%}&{59.70\%}&{56.79\%}&{64.44\%}&{55.71\%}&{68.69\%}\\
{5\%}&{62.10\%}&{65.77\%}&{53.54\%}&{62.54\%}&{61.00\%}&{68.39\%}&{60.70\%}&{75.24\%}\\
\hline
\end{tabular}
\caption{Protein}\label{tab:Protein}
\end{table*}

\begin{table*}[!]
\scriptsize
\centering
\begin{tabular}{|c|cccccc|cc|}
\hline
{}&\multicolumn{6}{c|}{Low rank approximation}&\multicolumn{2}{c|}{Raw Data}\\
{}&\multicolumn{2}{c}{GLNP}&\multicolumn{2}{c}{Nystr\"{o}m (random)}&\multicolumn{2}{c|}{Nystr\"{o}m (k-means)}&\multirow{ 2}{*}{KNN}&\multirow{2}{*}{LP (RBF)}\\
{\% of training}&{k=10}&{k=100}&{k=10}&{k=100}&{k=10}&{k=100}&&\\
\hline
{0.1\%}&{74.60\%}&{77.10\%}&{68.63\%}&{73.24\%}&{76.81\%}&{74.73\%}&{63.22\%}&{74.77\%}\\
{0.5\%}&{84.53\%}&{82.60\%}&{69.31\%}&{79.09\%}&{82.07\%}&{82.34\%}&{79.02\%}&{83.50\%}\\
{1\%}&{87.06\%}&{86.56\%}&{70.17\%}&{81.10\%}&{83.23\%}&{85.01\%}&{84.18\%}&{86.09\%}\\
{5\%}&{87.83\%}&{95.01\%}&{70.17\%}&{80.80\%}&{83.47\%}&{85.29\%}&{90.89\%}&{87.10\%}\\
\hline
\end{tabular}
\caption{Gisette}\label{tab:Gisette}
\end{table*}